\documentclass{article}
\usepackage{spconf,amsmath,graphicx}
\usepackage{booktabs} 
\usepackage{stfloats} 
\usepackage{multirow} 
\usepackage{color}
\newcommand{\ieno}{\textit{i.e.}}
\newcommand{\egno}{\textit{e.g.}}
\usepackage{amsmath}
\usepackage{graphicx}
\usepackage{epstopdf}
\usepackage{subfigure}
\usepackage{setspace}
\usepackage{times}

\title{PRIORFORMER: A UGC-VQA METHOD WITH CONTENT AND DISTORTION PRIORS}
\name{Yajing Pei$^*$  \qquad Shiyu Huang$^*$ \qquad Yiting Lu \qquad Xin Li$^{\dagger}$ \qquad Zhibo Chen$^{\dagger}$
\thanks{\hspace{-5mm} $^*$ Equal Contribution 
 \quad \quad $^{\dagger}$ Corresponding authors.}}
\address{University of Science and Technology of China, Hefei, China\\
\small \{peiyj,priest23,luyt31415,lixin666\}@mail.ustc.edu.cn, {chenzhibo@ustc.edu.cn}}

\begin{document}
%
\maketitle
\begin{abstract}
User Generated Content (UGC) videos are susceptible to complicated and variant degradations and contents, which prevents the existing blind video quality assessment (BVQA) models from good performance since the lack of the adaptability of distortions and contents. To mitigate this, we propose a novel prior-augmented perceptual vision transformer (PriorFormer) for the BVQA of UGC, which boots its adaptability and representation capability for divergent contents and distortions. Concretely, we introduce two powerful priors, \ieno, the content and distortion priors, by extracting the content and distortion embeddings from two pre-trained feature extractors. Then we adopt these two powerful embeddings as the adaptive prior tokens, which are transferred to the vision transformer backbone jointly with implicit quality features. Based on the above strategy, the proposed PriorFormer achieves state-of-the-art performance on three public UGC VQA datasets including KoNViD-1K, LIVE-VQC and YouTube-UGC.  
\end{abstract}
\section{Introduction}
\label{sec:intro}
Millions of User Generated Content (UGC) videos have sprung up with the rapid development of multi-media and mobile camera technologies, which brings a crucial and promising challenge, \ieno, how to measure the subjective quality of UGC videos accurately and properly. Different from other video datasets, UGC videos are usually acquisited and uploaded by amateur photographers.Therefore, the UGC videos are susceptible to  extremely diverse and complicated degradations, \ieno, the hybrid distortions, including  underexposure,overexposure, jitter, noise, color shift, etc. Apart from this, the content of UGC videos is generally very diverse due to the low requirements for shooting locations, such as natural scenes, animations, games, screen content, etc. These two aspects (complicated distortions and diverse contents) severely hinder the application of existing video quality assessment (VQA) methods to UGC videos. It is urgent to investigate an effective UGC VQA method to overcome this challenge and achieve human-like quality assessment for UGC videos.

Existing VQA methods can be roughly divided into three categories based on whether the original reference is required, including Full Reference (FR), Reduced Reference (RR), and Non Reference (NR). Among them, NR-VQA is a more challenging and commonly investigated task since no reference video is provided in most scenarios.
Traditional NR-VQA methods tend to exploit hand-crafted features~\cite{culibrk2009feature}, such as statistics or kernel functions, to measure the video quality. In recent years, we have witnessed the great development of deep neural networks (DNNs) for the VQA of UGC videos. For instance, RAPIQUE \cite{tu2021rapique} improves VQA with a combination of spatio-temporal statistics and learned features. In VSFA \cite{li2019quality}, a pre-trained CNN network is used to extract spatial features and two crucial effects of HVS (\egno, content-dependency, and temporal-memory effects) are incorporated to obtain temporal features.

Thanks to the development of transformer in natural language processing (NLP), most works move a step forward to explore the vision transformer for computer vision tasks, revealing the strong representational capability of the transformer. Unlike CNNs, the commonly-used vision transformer takes advantage of multi-head self-attention to model the long-range dependencies that exist in multiple image tokens. Recently, some attempts with the vision transformer have been made in image/video quality assessment~\cite{you2021transformer, golestaneh2022no, ke2021musiq,lu2024kvq,li2024ntire,liu2022swiniqa}.
You \textit{et al.} \cite{you2021transformer} utilizes a shallow Transformer encoder to capture the interaction between features of multiple image patches to modeling the interaction of quality information over the whole image. TReS~\cite{golestaneh2022no} proposes a hybrid combination of CNNs and Transformers features to extract an enhanced quality-aware representation. MUSIQ~\cite{ke2021musiq} designed a multi-scale Transformer with scale-based position embedding which can handle full-size image input with varying resolutions and aspect ratios.
However, the above works ignore the intrinsic characteristics of the contents and distortions existing in the UGC videos, which limits their application in real scenarios.

It is noteworthy that the perception is commonly determined by the distortions and contents in UGC videos. However, simply optimizing the VQA model with collected UGC datasets only provides the relation between UGC videos and their quality scores, which lacks explicit perception of distortions and contents. This leads to the weak adaptability of VQA models for the divergent contents and distortions and results in bad performance in UGC VQA. To improve the adaptability and representation ability of the VQA model, we propose the simple but effective prior-augmented vision transformer (PriorFormer), where the transformer is enabled by the content-distortion priors to better serve for UGC VQA. Particularly, we aim to incorporate the explicit content embedding and distortion embedding extracted from UGC videos into the Vision Transformer, where the content and distortion priors can increase the adaptability of the transformer to complicated and variant distortions and contents.  
Concretely, one pretrained large vision model (LVM) is explored to extract the abundant prior of contents/semantics. Following the GraphIQA~\cite{sun2022graphiqa}, we utilize a pretrained distortion graph  with multiple distortion types and levels as the distortion prior extractor. Then these two powerful priors are transferred to the vision transformer backbone to enable the transformer to obtain a more accurate quality assessment for each frame. To further fuse the frame-level qualities, we follow VSFA~\cite{li2019quality} and exploit the gated recurrent unit (GRU) and subjectively-inspired temporal pooling layer.
We evaluate our PriorFormer on three typical UGC datasets, including KoNViD-1k~\cite{hosu2017konstanz}, LIVE-VQC~\cite{sinno2018large} and Youtube-UGC~\cite{wang2019youtube}, of which the experiments demonstrate the effectiveness of our powerful PriorFormer. 


\begin{itemize}
    \item We argue that the adaptability of contents and distortions is crucial for the UGC VQA model, and the complicated and divergent contents and distortions of UGC video hinder the good performance of existing VQA methods. 
    
    
    \item We propose the novel PriorFormer for UGC-VQA, which is enabled by the powerful content and distortion priors to have excellent adaptability and performance for UGC videos. Particularly, we succeed in  extracting fine-grained content prior with a large vision-language model CLIP, and distortion priors by establishing the distortion graph.
    
    \item The proposed model achieves state-of-the-art performance on the three popular UGC VQA databases. Extensive experiments and ablation studies prove the effectiveness of our method.
\end{itemize}

\section{RELATED WORK}
\label{related work}
In the field of User-Generated Content Video Quality Assessment (UGC-VQA) (\ieno, Non-Reference Video Quality Assessment ), two primary research trajectories have emerged~\cite{V-BLINDS,simpleVQA,Dover}: traditional methods, and deep learning-based methods. The conventional approach, often limited by the constraints of handcrafted features and a lack of adaptability to complex UGC datasets~\cite{V-BLINDS}.
With the advancement of deep learning, the learning-based methods boost the performance of model on UGC-VQA tasks, which are specially designed for adaptive spatial-temporal fusion to handle spatio-temporal quality variation~\cite{VSFA,discovqa,simpleVQA}. VSFA~\cite{VSFA} employs GRU module to model the quality fusion along the time-domain. Considering the content importance of each frame to the global content of the full video, DiscoVQA~\cite{discovqa} design a content-aware attention mechanism to model the context relationship of multiple frames for temporal quality fusion. Yun \textit{et.al} conduct sparse frame sampling in the multi-path temporal module to model the quality aware temporal interaction. FastVQA~\cite{FastVQA, Dover} find the fragment-level input is enough for capturing distortion in spatio-temporal domain, which can eliminate substantial spatio-temporal redundancies. 
However, the above works ignore the intrinsic characteristics of the contents and distortions existing in the UGC videos, which limits their application in real scenarios.

\section{PROPOSED METHOD}
\label{sec:majhead}
\begin{figure*}
	\centering
	\includegraphics[width=\linewidth]{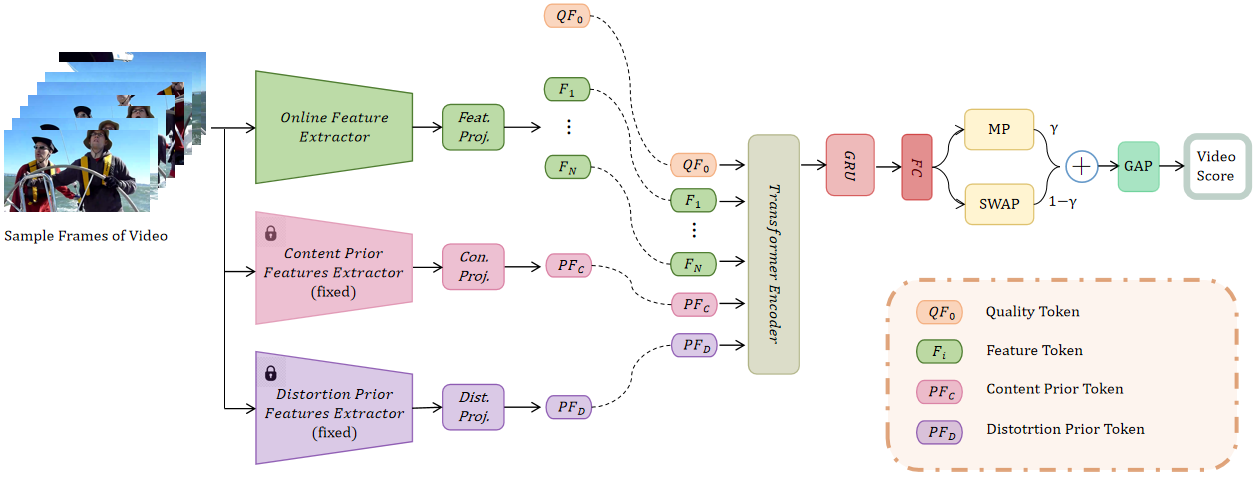}
	\caption{The overall framework of the proposed method. Frame features are extracted by an online feature extractor, and content and distortion prior features are extracted by fixed pretrained content and distortion prior feature extractors respectively. The extracted features are projected accordingly and inputted into the Transformer encoder to extract spatial features of the frames. Then, the GRU network and subjectively inspired temporal pooling layer(\ieno, Minimal Pooling, MP and Softmin-Weighted Average Pooling, SWAP) are utilized to model temporal-memory effects.}
	\label{fig:fw}
\end{figure*}

We first introduce the motivation of importing the content and distortion priors and the connections between them and UGC video quality. Then we clarify how to extract the distortion and content priors and incorporate the priors into the vision transformer architecture. Finally, we describe the spatial feature extraction based on Vision Transformer and temporal feature fusion modules. The overall framework of our proposed PriorFormer is illustrated in Fig.~\ref{fig:fw},
consisting of an online feature extractor, content and distortion prior feature extractors, prior-augmented Transformer encoder and a temporal feature fusion module.

\subsection{Motivation}
\label{sec:subhead}

Different from regular distortion video databases such as LIVE-VQA \cite{seshadrinathan2010study}, which are degraded by only several synthetic distortions and limited scene contents, the UGC videos usually have complex and divergent real-world contents and distortions. 

The perceptual quality of UGC VQA is consistently associated with the contents and distortions. As illustrated in YT-UGC+ , there is a strong correlation between MOS and corresponding content labels of UGC videos.
The reason for that is the human tolerance levels for the same distortion will change with different video contents.
Thus a good UGC quality metric should have adaptability to different contents.
Meanwhile, a good distortion representation is crucial for the success of deep blind image/video quality assessment. GraphIQA \cite{sun2022graphiqa} investigated how perceptual quality is affected by distortion-related factors and found the samples with the same distortion and level tend to have similar characteristics of distortion. 

\begin{figure}[ht] 
\centering 
\includegraphics[width=0.47\textwidth]{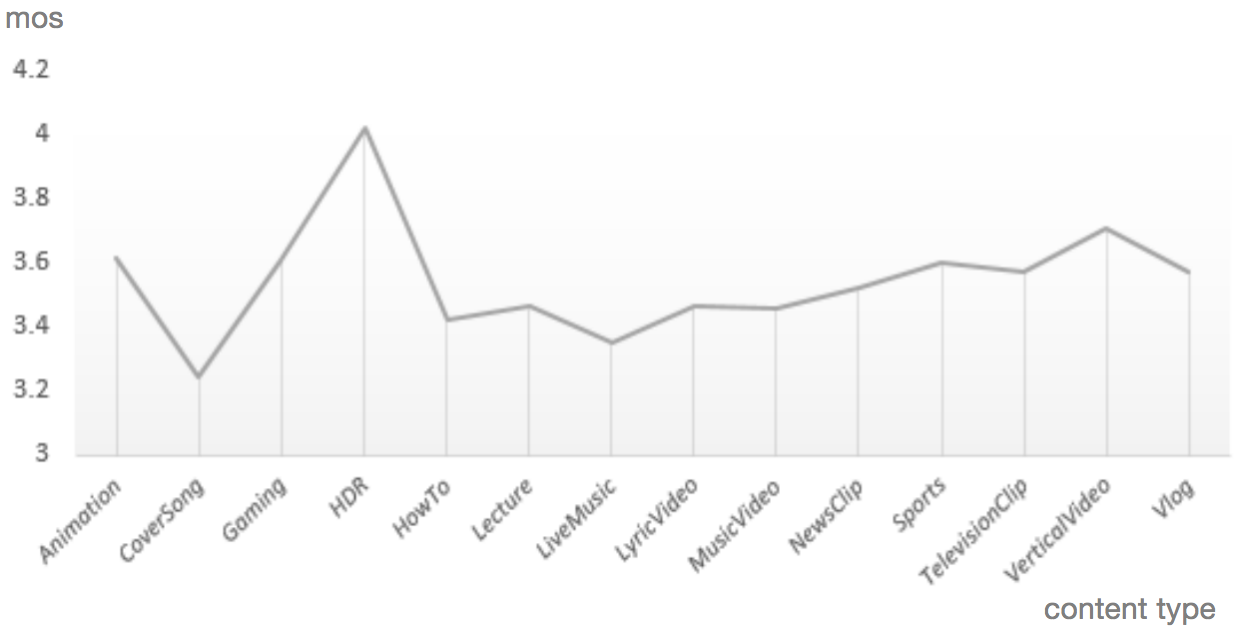} 
\caption{The MOS distribution of different content.} 
\label{fig:con} 
\end{figure}

\textbf{The preliminary analysis of content's impacts on quality.} In our study of the YouTube-UGC dataset~\cite{youtubeUGC}, we observed its extensive content diversity, encompassing 1500 video sequences, each 20 seconds in length and spanning across 15 distinct content categories. The analysis in Fig.~\ref{fig:con} shows the distribution of Mean Opinion Scores (MOS) across each content category within the YT-UGC dataset. Notably, HDR content exhibits the highest average quality (4.02), while CoverSong is at the lower end of the quality value (3.25). Categories like Gaming, Sports, and VerticalVideo tend to align with higher quality ranges. However, the substantial standard deviation of MOS across all content categories indicates the challenge in mapping high-level content labels to the perceived quality of videos, which underscores the significant role of video content in quality assessment. 

\begin{figure}[ht] 
\centering 
\includegraphics[width=0.5\textwidth]{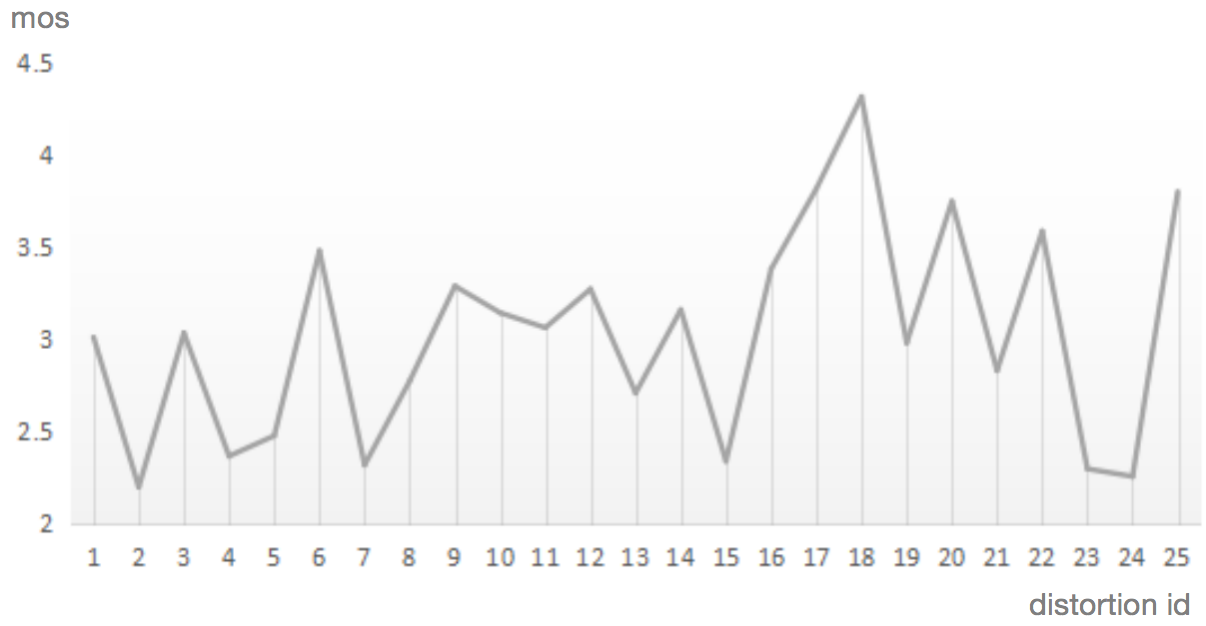} 
\caption{The MOS distribution of different distortion type.} 
\label{fig:dis} 
\end{figure}

\textbf{The preliminary analysis of distortion's impacts on quality.} Similarly, a robust representation of distortions is crucial for non-reference video quality assessment. To further investigate the impact of distortion-related factors on perceived quality, we conducted a study based on the Kadid-10k dataset ~\cite{lin2019kadid}, a large-scale collection comprising 10225 images, 25 types of distortions, and 5 distortion levels. The distribution of MOS for the 25 distortion types in the Kadid-10k dataset is depicted in Fig~\ref{fig:dis} . The substantial standard deviation of MOS in different distortion types highlights the significant role of distortion type in determining quality scores. 

Motivated by these, we aim to leverage the content and distortion priors to optimize the UGC VQA task. Inspired by the attention module in Transformer architecture, we consider incorporating the content and distortion priors of UGC into the Vision Transformer architecture through self-attention, which helps our model achieve better performance on UGC VQA.

\subsection{Prior-augmented Vision Transformer}
\subsubsection{Content and Distortion Prior Features Extractor}
\label{ssec:subsubhead}

A more fine-grained content prior is crucial for increasing the adaptability of the VQA model. To achieve this, we exploit a large pre-trained vision language model CLIP~\cite{radford2021learning} to extract the content prior. Particularly, the CLIP is pre-trained with large vision-language datasets of 400 million pairs. The cross-modality contrastive learning empowers the fine-grained content understanding of CLIP.
Therefore, we choose the image encoder ViT-B/16 in CLIP models as our Content Prior Features Extractor. Since these two pre-trained feature extraction structures have been introduced in the past work, we do not show the specific architecture.


Another challenge is how to extract the distortion prior effectively. To achieve this,  we utilize the distortion graph representation (DRG) in GraphIQA~\cite{sun2022graphiqa} to extract the distortion prior. In particular, DGR is pretrained on Kadid-10k \cite{lin2019kadid} and Kadis-700k \cite{lin2019kadid} to build the distortion graph for each specific distortion, and the DGR can be utilized to infer distortion type and level based on its internal structure.
Consequently, the DGR is able to extract plentiful distortion priors, which is proper to be applied as the distortion prior feature extractor.

\subsubsection{Prior-augmented Transformer Encoder}
\label{ssssec:subhead}

In order to leverage the content and distortion priors more efficiently, we propose a prior-augmented Transformer encoder, which incorporates the content and distortion priors into the transformer through self-attention. Concretely, we utilize
 the ResNet-50 as the online feature extractor to extract the quality-related features. Then the quality-related feature is
 projected to the fixed size of vectors and flattened to feature tokens ${F} \in {R} ^{ {N} \times {C}}$, where ${C}$ is the number of channels, ${N}=  {HW}/ {P^{2}}$ is the resulting number of feature tokens, and $\left ( \frac{H}{P},\frac{W}{P}\right )$ is the resolution of each feature map. Second, the content and distortion prior features extracted by two fixed Prior Features Extractors are projected to content/distortion prior tokens ($ {PF_{C} }  $, $ {PF_{D} }  $). The trainable  quality token ($ {QF_{0} }  $) is then added to the sequence of embedded features to predict perceived quality. In order to maintain the positional information, the learnable position embeddings ($ {PE}  $) are added to the tokens. Subsequently, the quality token, feature tokens, and prior tokens are fed into a Transformer encoder. The attention layer in the Transformer encoder consists of multi-head attention (MHAs), position-wise feed-forward layers (FFs), layer normalizations (LNs), and residual connections. The calculation of the encoder can be formulated as:

    \begin{align}
         {Z}_{0} &= [{F}_{0} + {PE}_{0};{F}_{1:N} + {PE}_{1:N};{PF}_{C} + {PE}_{N+1};{PF}_{D} \notag\\ & +{PE}_{N+2}], 
          {PE_{j} \in{R}^{( {N+3} )\times{C}}}, \notag\\
        {Z}_{l}^{'} &={LN}({MHA} ({Z}_{l-1}) +{Z}_{l-1}), \\ \notag
        {Z}_{l} &= {LN}({FF}({Z}_{l}^{'}) +{Z}_{l}^{'} ),
        l =1,\cdots {L}, 
    \end{align}

Finally, the output embedding $Z_{L}^{0}$ corresponding to the quality token which claims to contain learned image quality information from the Transformer encoder, is extracted as the frame quality expression.

\subsubsection{Temporal Feature Fusion}
\label{sssssec:subhead}

 Inspired by VSFA\cite{li2019quality}, we utilize the GRU network to learn long-term dependencies between frames in the temporal domain, and the subjectively-inspired temporal pooling to model temporal memory effect, \ieno, the individuals responded quickly to video quality drop but slowly to video quality increase. 

The existing methods for BVQA fail to adequately model the long-term temporal dependencies in VQA tasks. To address this issue, we propose the utilization of a recurrent neural network, specifically a Gated Recurrent Unit (GRU), equipped with gating units, enabling it to fuse context-aware features, learn long-term dependencies across time sequences, and predict frame-level quality scores. Initially, $F_t$ is the quality representation of the $t-th$ frame, $F_t=Z_L^0$, $Z_L^0$ is the output of the $L-th$ layer of the Transformer encoder, which serves as the model's input. Subsequently, these features are fed into the GRU network. The hidden state of the GRU is perceived as an integrated feature, with an initial value of $h_0$, while the current hidden state $h_t$ at time $t$ is jointly computed based on the current spatial feature $F_t$ and the previous hidden state $h_{t-1}$. 
\begin{equation}\label{eq1}
h_t=GRU(F_t,h_{t-1})
\end{equation}
The learned integrated feature $h_t$ is then processed through a fully connected layer to yield the predicted single-frame quality score $q_t$. 
\begin{equation}\label{eq3}
q_t=W_{fc}h_t+b_{fc}
\end{equation}
Here, $W_{fc}$ and $b_{fc}$ denote the weights and bias parameters of the fully connected layer, respectively.

To simulate this temporal perception mechanism of the human visual system, we introduce a subjective-inspired temporal pooling layer model following ~\cite{5946613}. To emulate the low tolerance of the human eye towards low-quality video content, we define a memory quality element $m_t$ for the $t-th$ frame, representing the minimum quality score within the preceding $\tau$ duration.Where
\begin{equation}\label{eq2}
    \left\{
        \begin{array}{ll}
            m_t =q_t, & for \; t = 1 \\
            m_t = min_{k\in V_{prev}}q_t, & for \; t > 1.
        \end{array}
    \right.
\end{equation}

To emulate the subjects' reaction speed to video quality decline, we define a current quality element $c_t$ for the $t-th$ frame. It is represented by the weighted quality scores of frames within the $\tau$ time window, with poorer quality frames assigned greater weight. 


\begin{table*}[tp]
 
  \centering
  \caption{\bf Performance of the SOTA models and the proposed model on the KoNViD-1k, LIVE-VQC, and YouTube-UGC databases. }
  \vspace{+1em}
  \label{tab:performance_comparison}
    \begin{tabular}{l c c c c c c c c c}
    \hline
    \multirow{2}{*}{Method}&
    \multicolumn{2}{c}{KoNViD-1k}&\multicolumn{2}{c}{LIVE-VQC}&\multicolumn{2}{c}{YouTube-UGC}\cr\cline{2-3}\cline{4-5}\cline{6-7}
    &PLCC&SRCC&PLCC&SRCC&PLCC&SRCC\cr
    \hline
    NIQE \cite{mittal2012making}&0.5513&0.5392&0.6312&0.5930&0.2982&0.2499\cr
    BRISQUE \cite{mittal2012no}&0.6513&0.6493&0.6242&0.5936&0.4073&0.3932\cr
    VIIDEO \cite{mittal2015completely}&0.3083&0.2874&0.2100&0.0461&0.1497&0.0567\cr\hline
    TLVQM \cite{korhonen2019two}&0.7598&0.7588&0.7942&0.7878&0.6470&0.6568\cr
    VIDEVAL \cite{tu2021ugc}&0.7709&0.7704&0.7476&0.7438&0.7715&0.7763\cr
    VSFA \cite{li2019quality}&0.7958&0.7943&0.7707&0.7176&0.7888&0.7873\cr
    RAPIQUE \cite{tu2021rapique}&0.8175&0.8031&0.7863&0.7548&0.7684&0.7591\cr\hline
    ResNet-50 \cite{he2016deep}&0.7781&0.7651&0.7381&0.6814&0.6485&0.6542\cr
    CLIP \cite{radford2021learning}&0.7806&0.7892&0.7493&0.7235&0.7640&0.7771\cr
    GraphIQA \cite{sun2022graphiqa}&0.7667&0.7644&0.7146&0.6609&0.6687&0.6574\cr
    PaQ2PiQ \cite{PaQ2PiQ}&0.6014&0.6130&0.6683&0.6436&0.0.2935&0.2658\cr
    PVQ \cite{Patch-vq}&0.795&0.791&0.807&0.770&--&--\cr\hline
    
    Proposed&{\bf 0.8390}&{\bf 0.8291}&{\bf 0.8228}&{\bf 0.7966}&{\bf 0.8399}&{\bf 0.8394}\cr
    
    \hline
    \end{tabular}
\end{table*}

\section{EXPERIMENTS}
\label{sec:exp}

\subsection{Experimental Settings}
\label{ssssssec:subhead}

\textbf{Databases.} Our method is evaluated on 3 UGC-VQA databases to verify the effect: KoNViD-1K~\cite{hosu2017konstanz}, LIVE-VQC~\cite{sinno2018large}, and YouTube-UGC~\cite{wang2019youtube}. 

In detail, KoNViD-1k~\cite{hosu2017konstanz} contains 1,200 videos with 8 seconds and a resolution of 540p. The Mean Opinion Score (MOS) ranges from 1.22 to 4.64. LIVE-VQC~\cite{sinno2018large} consists of 585 videos with a duration of 10 seconds and resolutions from 240p to 1080p, captured by 80 different users with 101 different devices. The MOS of these videos ranges from 16.5621 to 73.6428. YouTube-UGC~\cite{wang2019youtube} comprises 1,380 UGC videos with 20 seconds, which is sampled from YouTube with resolutions from 360p to 4k and MOS ranging from 1.242 to 4.698. 
None of these datasets contain pristine videos, and we randomly split these datasets into 80$\%$  for the training set and 20$\%$ for the testing set.

\textbf{Evaluation criteria.} Two evaluation criteria are selected to evaluate the performance of our method, consisting of Pearson Linear Correlation Coefficient (PLCC) and Spearman Rank Order Correlation Coefficient (SROCC). The PLCC evaluates the linear relationship between predicted score and MOS value, while the SROCC measures the monotonicity. The value of these criteria is closer to 1, which means the correlation between the predicted score and MOS value is higher.

\textbf{Implementation details.} We implement our model by PyTorch, and both training and testing are conducted on the NVIDIA 1080Ti GPUs. The video is sampled at 1 frame per second(fps) and resized to 224$\times $224. The feature extracted by ResNet-50, content and distortion prior features are respectively projected and flattened to the 512-dimension vector as feature tokens, content prior tokens and distortion prior tokens. For the Transformer encoder, the hyper-parameters are set as: L(number of layers)=6, H(number of the heads in the MHA)=8, D(the Transformer dimension)=512, $\mathrm {D}_{ff} $(dimension of the feed-forward network)=1024. For the temporal feature extraction, the GRU hidden layer is set to 32, and $\gamma $ is set to 0.5. 

\subsection{Performance Comparison}
\label{sssssssec:subhead}

We select some representative BIQA/BVQA methods for comparison on three above datasets, including traditional methods based on hand-craft feature (\egno, NIQE \cite{mittal2012making}, BRISQ\\-UE \cite{mittal2012no}, and VIIDEO \cite{mittal2015completely}) and deep learning-based methods with well-designed networks (\egno, TLVQM \cite{korhonen2019two}, VIDEVAL \cite{tu2021ugc}, VSFA \cite{li2019quality}, and RAPIQUE \cite{tu2021rapique}). 
We also compare the performance of models with the same architecture as some modules in our proposed method:  ResNet-50 \cite{he2016deep}, the image encoder ViT-B/16 of CLIP \cite{radford2021learning}, and GraphIQA \cite{sun2022graphiqa}. For the BIQA method, we calculated the quality score of the entire video using the mean of the frame quality scores. The test results of each method are shown in TABLE~\ref{tab:performance_comparison}. It can be seen from the results that the proposed PriorFormer algorithm shows competitive performance on the three data sets.


\subsection{Ablation Study}

\begin{table*}[tp]
 
  \centering
  \caption{\bf Ablation Study of the proposed model on the KoNViD-1k, LIVE-VQC, and YouTube-UGC databases. }
  \vspace{+1em}
  \label{tab:Ablation Study}
    \begin{tabular}{l c c c c c c c c c}
    \hline
    \multirow{2}{*}{Method}&
    \multicolumn{2}{c}{KoNViD-1k}&\multicolumn{2}{c}{LIVE-VQC}&\multicolumn{2}{c}{YouTube-UGC}\cr\cline{2-3}\cline{4-5}\cline{6-7}
    &PLCC&SRCC&PLCC&SRCC&PLCC&SRCC\cr
    \hline
    PriorFormer &{\bf 0.8390}&{\bf 0.8291}&0.8228&{\bf 0.7966}&{\bf 0.8399}&{\bf 0.8394}\cr
    PriorFormer(w.o. CT) &0.8129&0.8171&0.7949&0.7476&0.7952&0.7905\cr
    PriorFormer(w.o. DT) &0.8193&0.8256&{\bf 0.8246}&0.7909&0.8353&0.8320\cr
    PriorFormer(w.o. CT+DT) &0.7785&0.7704&0.7546&0.7204&0.7861&0.7969\cr
    PriorFormer(w.o. TP)&0.8027&0.7994&0.7983&0.7796&0.8062&0.7985\cr
    
    \hline
    \end{tabular}
\end{table*}

To validate the importance of each module in the model, we conducted ablation experiments to assess the significance of the content and distortion priors, as well as the temporal perception mechanism in modeling. The results of the ablation experiments are presented in TABLE~\ref{tab:Ablation Study}. 

We first individually removed the content prior feature extractor and the distortion prior feature extractor, and then removed both simultaneously to evaluate the importance of both priors. It can be observed that removing either the content prior (w.o. CT) or the distortion prior (w.o. DT) leads to a decrease in model performance. Specifically, for the LIVE-VQC and YouTube-UGC datasets, the impact on performance by solely removing the distortion prior (w.o. DT) is relatively minor. This might be attributed to the fact that the UGC video dataset is influenced by diverse and complex real-world distortions, leading to quality degradation, whereas our distortion prior feature extraction sub-network is trained on synthetic distortion datasets, thus having limited capability to learn various distortion types and levels. However, for all tested datasets, removing the content prior (w.o. CT) results in a significant decrease in model performance, further indicating that semantic content significantly influences the perceived quality assessment of UGC videos. Moreover, compared to the scenario of removing both distortion and content priors simultaneously (w.o. CT+DT), it is evident that the combination of both priors enhances the model's performance, highlighting the close relationship between quality perception and content/distortion. 

 Additionally, we conducted experiments by only removing the temporal feature extraction module (w.o. TP), and the observed performance decline underscores the importance of the temporal perception mechanism.

\section{CONCLUSION}
\label{sec:conclusion}

In this paper, we propose a UGC VQA model based on prior-augmented perceptual vision transformer and temporal perception mechanisms. By incorporating content and distortion priors, our model meets the heightened requirements of UGC VQA for content and quality comprehension. Furthermore, through modeling the temporal perception mechanism, our model adeptly integrates temporal information, providing objective scores highly consistent with human perception. Our approach demonstrates outstanding performance across three public UGC VQA databases, and ablation experiments show the effectiveness of the content and distortion priors, as well as the temporal pooling model. Future work that focuses on utilizing more UGC-related prior knowledge is desirable.

\section{ACKNOWLEDGEMENTS}
\label{sec:acknowledgements}

This work was supported in part by NSFC under Grant 62371434, 62021001, and 623B2098. 


\bibliographystyle{IEEEbib}
 \bibliography{bibfile}   

\begin{thebibliography}{10}

\bibitem{culibrk2009feature}
Dubravko {\'C}ulibrk, Dragan Kukolj, Petar Vasiljevi{\'c}, Maja Pokri{\'c}, and Vladimir Zlokolica,
\newblock ``Feature selection for neural-network based no-reference video quality assessment,''
\newblock in {\em ICANN}. Springer, 2009, pp. 633--642.

\bibitem{tu2021rapique}
Zhengzhong Tu, Xiangxu Yu, Yilin Wang, Neil Birkbeck, Balu Adsumilli, and Alan~C Bovik,
\newblock ``Rapique: Rapid and accurate video quality prediction of user generated content,''
\newblock {\em IEEE Open Journal of Signal Processing}, vol. 2, pp. 425--440, 2021.

\bibitem{li2019quality}
Dingquan Li, Tingting Jiang, and Ming Jiang,
\newblock ``Quality assessment of in-the-wild videos,''
\newblock in {\em ACM MM}, 2019, pp. 2351--2359.

\bibitem{you2021transformer}
Junyong You and Jari Korhonen,
\newblock ``Transformer for image quality assessment,''
\newblock in {\em IEEE ICIP}, 2021, pp. 1389--1393.

\bibitem{golestaneh2022no}
S~Alireza Golestaneh, Saba Dadsetan, and Kris~M Kitani,
\newblock ``No-reference image quality assessment via transformers, relative ranking, and self-consistency,''
\newblock in {\em IEEE WACV}, 2022, pp. 1220--1230.

\bibitem{ke2021musiq}
Junjie Ke, Qifei Wang, Yilin Wang, Peyman Milanfar, and Feng Yang,
\newblock ``Musiq: Multi-scale image quality transformer,''
\newblock in {\em IEEE/CVF,ICCV}, 2021, pp. 5148--5157.

\bibitem{lu2024kvq}
Yiting Lu, Xin Li, Yajing Pei, Kun Yuan, Qizhi Xie, Yunpeng Qu, Ming Sun, Chao Zhou, and Zhibo Chen,
\newblock ``Kvq: Kwai video quality assessment for short-form videos,''
\newblock in {\em IEEE/CVF,CVPR}, 2024, pp. 25963--25973.

\bibitem{li2024ntire}
Xin Li, Kun Yuan, Yajing Pei, Yiting Lu, Ming Sun, Chao Zhou, Zhibo Chen, Radu Timofte, et~al.,
\newblock ``Ntire 2024 challenge on short-form ugc video quality assessment: Methods and results,''
\newblock {\em arXiv preprint arXiv:2404.11313}, 2024.

\bibitem{liu2022swiniqa}
Jianzhao Liu, Xin Li, Yanding Peng, Tao Yu, and Zhibo Chen,
\newblock ``Swiniqa: Learned swin distance for compressed image quality assessment,''
\newblock in {\em IEEE/CVF,CVPR}, 2022, pp. 1795--1799.

\bibitem{sun2022graphiqa}
Simeng Sun, Tao Yu, Jiahua Xu, Wei Zhou, and Zhibo Chen,
\newblock ``Graphiqa: Learning distortion graph representations for blind image quality assessment,''
\newblock {\em IEEE TMM}, 2022.

\bibitem{hosu2017konstanz}
Vlad Hosu, Franz Hahn, Mohsen Jenadeleh, Hanhe Lin, Hui Men, Tam{\'a}s Szir{\'a}nyi, Shujun Li, and Dietmar Saupe,
\newblock ``The konstanz natural video database (konvid-1k),''
\newblock in {\em IEEE QoMEX}, 2017, pp. 1--6.

\bibitem{sinno2018large}
Zeina Sinno and Alan~Conrad Bovik,
\newblock ``Large-scale study of perceptual video quality,''
\newblock {\em IEEE TIP}, vol. 28, no. 2, pp. 612--627, 2018.

\bibitem{wang2019youtube}
Yilin Wang, Sasi Inguva, and Balu Adsumilli,
\newblock ``Youtube ugc dataset for video compression research,''
\newblock in {\em IEEE MMSP}, 2019, pp. 1--5.

\bibitem{V-BLINDS}
Michele~A. Saad, Alan~C. Bovik, and Christophe Charrier,
\newblock ``Blind prediction of natural video quality,''
\newblock {\em IEEE TIP}, vol. 23, no. 3, pp. 1352--1365, 2014.

\bibitem{simpleVQA}
Wei Sun, Xiongkuo Min, Wei Lu, and Guangtao Zhai,
\newblock ``A deep learning based no-reference quality assessment model for {UGC} videos,''
\newblock in {\em ACM MM}, 2022, pp. 856--865.

\bibitem{Dover}
Haoning Wu, Liang Liao, Chaofeng Chen, Jingwen Hou, Annan Wang, Wenxiu Sun, Qiong Yan, and Weisi Lin,
\newblock ``Disentangling aesthetic and technical effects for video quality assessment of user generated content,''
\newblock {\em CoRR}, vol. abs/2211.04894, 2022.

\bibitem{VSFA}
Dingquan Li, Tingting Jiang, and Ming Jiang,
\newblock ``Quality assessment of in-the-wild videos,''
\newblock in {\em ACM MM}, 2019, pp. 2351--2359.

\bibitem{discovqa}
Haoning Wu, Chaofeng Chen, Liang Liao, Jingwen Hou, Wenxiu Sun, Qiong Yan, and Weisi Lin,
\newblock ``Discovqa: Temporal distortion-content transformers for video quality assessment,''
\newblock {\em IEEE TCSVT}, vol. 33, no. 9, pp. 4840--4854, 2023.

\bibitem{FastVQA}
Haoning Wu, Chaofeng Chen, Jingwen Hou, Liang Liao, Annan Wang, Wenxiu Sun, Qiong Yan, and Weisi Lin,
\newblock ``{FAST-VQA:} efficient end-to-end video quality assessment with fragment sampling,''
\newblock in {\em {ECCV} {(6)}}. 2022, vol. 13666 of {\em Lecture Notes in Computer Science}, pp. 538--554, Springer.

\bibitem{seshadrinathan2010study}
Kalpana Seshadrinathan, Rajiv Soundararajan, Alan~Conrad Bovik, and Lawrence~K Cormack,
\newblock ``Study of subjective and objective quality assessment of video,''
\newblock {\em IEEE TIP}, vol. 19, no. 6, pp. 1427--1441, 2010.

\bibitem{youtubeUGC}
Yilin Wang, Sasi Inguva, and Balu Adsumilli,
\newblock ``Youtube {UGC} dataset for video compression research,''
\newblock in {\em {MMSP}}. 2019, pp. 1--5, {IEEE}.

\bibitem{lin2019kadid}
Hanhe Lin, Vlad Hosu, and Dietmar Saupe,
\newblock ``Kadid-10k: A large-scale artificially distorted iqa database,''
\newblock in {\em IEEE QoMEX}, 2019, pp. 1--3.

\bibitem{radford2021learning}
Alec Radford, Jong~Wook Kim, Chris Hallacy, Aditya Ramesh, Gabriel Goh, Sandhini Agarwal, Girish Sastry, Amanda Askell, Pamela Mishkin, Jack Clark, et~al.,
\newblock ``Learning transferable visual models from natural language supervision,''
\newblock in {\em International conference on machine learning}. PMLR, 2021, pp. 8748--8763.

\bibitem{5946613}
Kalpana Seshadrinathan and Alan~C. Bovik,
\newblock ``Temporal hysteresis model of time varying subjective video quality,''
\newblock in {\em IEEE ICASSP}, 2011, pp. 1153--1156.

\bibitem{mittal2012making}
Anish Mittal, Rajiv Soundararajan, and Alan~C Bovik,
\newblock ``Making a “completely blind” image quality analyzer,''
\newblock {\em IEEE Signal processing letters}, vol. 20, no. 3, pp. 209--212, 2012.

\bibitem{mittal2012no}
Anish Mittal, Anush~Krishna Moorthy, and Alan~Conrad Bovik,
\newblock ``No-reference image quality assessment in the spatial domain,''
\newblock {\em IEEE TIP}, vol. 21, no. 12, pp. 4695--4708, 2012.

\bibitem{mittal2015completely}
Anish Mittal, Michele~A Saad, and Alan~C Bovik,
\newblock ``A completely blind video integrity oracle,''
\newblock {\em IEEE TIP}, vol. 25, no. 1, pp. 289--300, 2015.

\bibitem{korhonen2019two}
Jari Korhonen,
\newblock ``Two-level approach for no-reference consumer video quality assessment,''
\newblock {\em IEEE TIP}, vol. 28, no. 12, pp. 5923--5938, 2019.

\bibitem{tu2021ugc}
Zhengzhong Tu, Yilin Wang, Neil Birkbeck, Balu Adsumilli, and Alan~C Bovik,
\newblock ``Ugc-vqa: Benchmarking blind video quality assessment for user generated content,''
\newblock {\em IEEE TIP}, vol. 30, pp. 4449--4464, 2021.

\bibitem{he2016deep}
Kaiming He, Xiangyu Zhang, Shaoqing Ren, and Jian Sun,
\newblock ``Deep residual learning for image recognition,''
\newblock in {\em IEEE/CVF,CVPR}, 2016, pp. 770--778.

\bibitem{PaQ2PiQ}
Ying Z, Niu H, and et~al Gupta~P,
\newblock ``From patches to pictures (paq-2-piq): Map** the perceptual space of picture quality[c],''
\newblock {\em IEEE/CVF CVPR. 2020: 3575-3585.}, 2020.

\bibitem{Patch-vq}
Ying Z, Mandal M, and et~al Ghadiyaram~D,
\newblock ``Patch-vq:'patching up'the video quality problem,''
\newblock {\em IEEE/CVF,CVPR. 2021: 14019-14029}, 2021.

\end{thebibliography}
\end{document}